\def\year{2017}\relax
\documentclass[letterpaper]{article}
\usepackage{aaai17}
\usepackage{times}
\usepackage{helvet}
\usepackage{courier}

\usepackage{url}            
\usepackage{booktabs}       
\usepackage{amsfonts}       
\usepackage{nicefrac}       
\usepackage{subfig}
\usepackage{graphicx}
\usepackage{multirow}
\usepackage{amsmath}
\usepackage{array}
\usepackage{amsmath}
\usepackage{amssymb}

\begin{document}
%
\title{Beyond Monte Carlo Tree Search: Playing Go with \\ Deep Alternative Neural Network and Long-Term Evaluation}
\author{Jinzhuo Wang, Wenmin Wang, Ronggang Wang, Wen Gao\textsuperscript{\textdagger}\\
School of Electronics and Computer Engineering, Peking University\\
\textsuperscript{\textdagger}School of Electronics Engineering and Computer Science, Peking University\\
jzwang@pku.edu.cn, wangwm@ece.pku.edu.cn, rgwang@ece.pku.edu.cn, wgao@pku.edu.cn\\
}
\maketitle

\begin{abstract}
Monte Carlo tree search (MCTS) is extremely popular in computer Go which determines each action by enormous simulations in a broad and deep search tree. However, human experts select most actions by pattern analysis and careful evaluation rather than brute search of millions of future interactions. In this paper, we propose a computer Go system that follows experts¡¯ way of thinking and playing. Our system
consists of two parts. The first part is a novel deep alternative neural network (DANN) used to generate candidates of next move. Compared with existing deep convolutional neural network (DCNN), DANN inserts recurrent layer after each convolutional layer and stacks them in an alternative manner. We show such setting can preserve more contexts of local features and its evolutions which are beneficial for move prediction. The second part is a long-term evaluation (LTE) module used to provide a reliable evaluation of candidates rather than a single probability from move predictor. This is consistent with human experts¡¯ nature of playing since they can foresee tens of steps to give an accurate estimation of candidates. In our system, for each candidate, LTE calculates a cumulative reward after several future interactions when local variations are settled. Combining criteria from the two parts, our system determines the optimal choice of next move. For more comprehensive experiments, we introduce a new professional Go dataset (PGD), consisting of $253,233$ professional records. Experiments on GoGoD and PGD datasets show the DANN can substantially improve performance of move prediction over pure DCNN. When combining LTE, our system outperforms
most relevant approaches and open engines based on MCTS.
\end{abstract}

\section{Introduction}

Go is a game of profound complexity and draws a lot attention. Although its rules are very simple \cite{muller2002computer}, it is difficult to construct a suitable value function of actions in most of situations mainly due to its high branching factors and subtle board situations that are sensitive to small changes.



Previous solutions focus on simulating future possible interactions to evaluate candidates. In such methods, Monte Carlo tree search (MCTS) \cite{gelly2011monte} is the most popular one, which constructs a broad and deep search tree to simulate and evaluate each action. However, the playing strength of MCTS-based Go programs is still far from human-level due to its major limitation of uneven performance. Well known weaknesses include {\it{capturing race}} or {\it{semeai}}, positions with multiple cumulative evaluation errors, {\it{ko fights}}, and close endgames \cite{rimmel2010current,huang2013investigating}. We attribute it to the following reasons. First, the effectivity of truncating search tree is based on prior knowledge and far away from perfect play \cite{muller2002computer}. Second, when the board is spacious especially at {\it{opening}}, simulation is expensive and useless. Besides, the outputs of leaves in Monte Carlo tree are difficult to be precisely evaluated \cite{browne2012survey}. Last but most important, MCTS does not follow professionals' way of playing since professionals hardly make brute simulation of every possible future positions. Instead, in most situations, they first obtain some candidates using pattern analysis and determine the optimal one by evaluating these candidates.

Recently as deep learning revolutionizes and gradually dominate many tasks in computer vision community, researcher start to borrow deep learning techniques for move prediction and develop computer Go systems \cite{clark2015training,maddison2015move,tian2016better,silver2016mastering}. However, compared with visual signals (e.g. $224 \times 224$ in image domain), Go board has a much smaller size ($19 \times 19$), which poses the importance of relative position. This is consistent with playing Go as situation can dramatically alter with a minor change in position. On the other hand, existing DCNNs often mine such contexts by stacking more convolutional layers (e.g. up to $13$ layers in \cite{silver2016mastering}) to exploit high-order encodings of low-level features. Simply increasing layers not only suffers parameter burden but also does not embed contexts of local features and its evolutions.

Based on the above discussions, this paper introduces a computer Go system consisting of two major parts. The first part is a novel deep architecture used to provide a probability distribution of legal candidates learned from professionals' records. These candidates are further sent to a long-term evaluation part by considering local future impact instead of immediate reward. We expect the model focus on several suggested important regions rather than blind simulation of every corner in the board. The primary contributions of this work are summarized as follows.
\begin{itemize}
    \item   We propose a novel deep alternative neural network (DANN) to learn a pattern recognizer for move prediction. The proposed DANN enjoys the advantages of both CNN and recurrent neural network (RNN), preserving contexts of local features and its evolutions which we show are essential for playing Go. Compared with existing DCNN-based models, DANN can substantially improve the move prediction performance using less layers and parameters.
    \item   To further enhance the candidates generated from DANN, we present a novel recurrent model to make a long-term evaluation around each candidate for the final choice. We formulate the process as a partially observe Markov decision process (POMDP) problem and propose a reinforcement learning solution with careful control of variance.
    \item   We introduce a new professional Go dataset (PGD) for comprehensive evaluation. PGD consists of $329.4$k modern professional records and considered useful for computer Go community. Thorough experiments on GoGoD and PGD demonstrates the advantages of our system over DCNNs and MCTS on both move prediction and win rate against open source engines.
\end{itemize}

\section{Related Work}


\textbf{Monte Carlo tree search (MCTS).} It is a best-first search method based on randomized explorations of search space, which does not require a positional evaluation function \cite{browne2012survey}. Using the results of previous explorations, the algorithm gradually grows a game tree, and successively becomes better at accurately estimating the values of the optimal moves \cite{bouzy2004monte} \cite{coulom2006efficient}. Such programs have led to strong amateur level performance, but a considerable gap still remains between top professionals and the strongest computer programs. The majority of recent progress is due to increased quantity and quality of prior knowledge, which is used to bias the search towards more promising states, and it is widely believed that this knowledge is the major bottleneck towards further progress. The first successful current Go program \cite{kocsis2006bandit} was based on MCTS. Their basic algorithm was augmented in MoGo \cite{gelly2007combining} to leverage prior knowledge to bootstrap value estimates in the search tree. Training with professionals' moves was enhanced in Fuego \cite{enzenberger2010fuego} and Pachi \cite{baudivs2011pachi} and achieved strong amateur level.

\textbf{Supervised pattern-matching policy learning.} Go professionals rely heavily on pattern analysis rather than brute simulation in most cases \cite{clark2015training,xiao2016factorization}. They can gain strong intuitions about what are the best moves to consider at a glance. This is in contrast to MCTS which simulates enormous possible future positions. The prediction functions are expected to be non-smooth and highly complex, since it is fair to assume professionals think in complex, non-linear ways when they choose moves. To this end, neural networks especially CNNs are widely used \cite{schraudolph1994temporal,enzenberger1996integration,richards1998evolving,sutskever2008mimicking}. Recent works in image recognition have demonstrated considerable advantages of DCNNs \cite{krizhevsky2012imagenet,simonyan2014very,szegedy2015going} and showed substantial improvement over shallow networks based on manually designed features or simple patterns extracted from previous games \cite{silver2009reinforcement}. DCNNs have yielded several state-of-the-art Go playing system \cite{clark2015training,maddison2015move,tian2016better,silver2016mastering} with $8,12$ and $13$ convolutional layers. Besides, these works have also indicated combining DCNNs and MCTS can improve the overall playing strength. Similar conclusions are validated in state-of-the-art Go programs \cite{enzenberger2010fuego,baudivs2011pachi}. The major difference between this paper and existing combinations comes from two perspectives. The first one is that we use a novel architecture DANN to generate candidates with more consideration of local contexts and its evolutions. The proposed architecture shows substantial improvement with pure DCNN with fewer layers and parameters. The second is that we use a long-term evaluation to analyze previous candidates instead of MCTS to assist the final choice. This strategy is faster than MCTS because the former needs to consider a large search space.


\section{The Proposed Computer Go System}
\begin{figure*}[t]
  \centering
  \includegraphics[width=17.6cm]{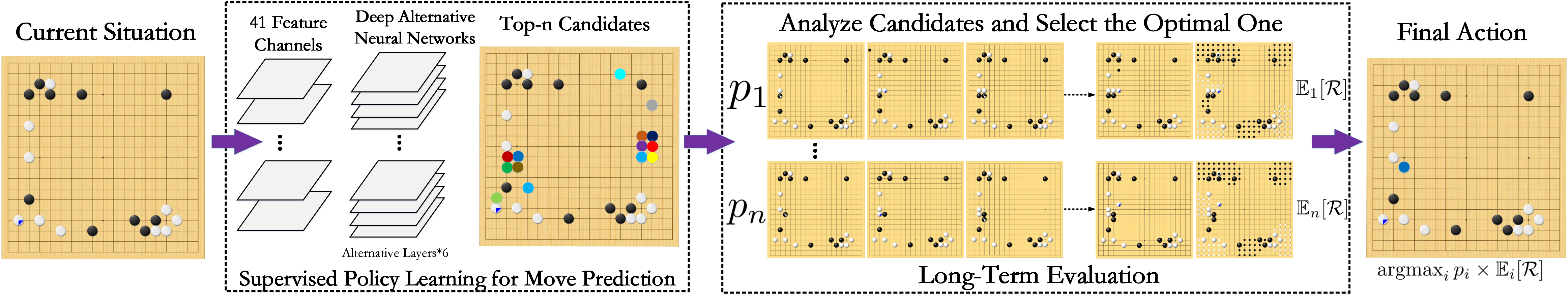}\label{fig:sys}
  \caption{The proposed computer Go system with deep alternative neural network (DANN) and long-term evaluation. Given a situation the system generate several candidates by DANN that are learned from professional records. These candidates are further analyzed using a long-term evaluation with consideration of future rewards to determine a final action.}\label{fig:sys}
\end{figure*}

The proposed system is illustrated in Figure \ref{fig:sys}). Given a situation, we first obtain a probability distribution of legal points that are learned from a pattern-aware prediction instrument based on supervised policy learning from existing professional records. We then further analyze those candidates with high-confidence by a long-term evaluation to pursue a expectation reward after several future steps when the local situation is settled. The action with the highest score of criteria combination of two criteria is our final choice.

\subsection{Deep Alternative Neural Network}

We train a novel deep alternative neural network (DANN) to generate a probability distribution of legal points given current board situation as an input. We treat the $19 \times 19$ board as a $19 \times 19$ image with multiple channels. Each channel encodes a different aspect of board information (see details in Table \ref{tab:feature}). In the following we describe the structure of DANN including its key component (alternative layer) and overall architecture. Afterwards we discuss its relations and advantages over popular DCNNs.

\textbf{Alternative layer.}\label{sec:dann:al} The key component of DANN is the alternative layer (AL), which consists of a standard convolutional layer followed by a designed recurrent layer. Specifically, convolution operation is first performed to extract features from local neighborhoods on feature maps in the previous layers. Then a recurrent layer is applied to the output and iteratively proceeds for $T$ times. This procedure makes each unit evolve over discrete time steps and aggregate larger receptive fields (RFs). More formally, the input of a unit at position $(x,y,z)$ in the $j$th feature map of the $i$th AL at time $t$, denoted as $u_{ij}^{xyz}(t)$, is given by
\begin{equation}\label{eq:al}
    \begin{split}
        u_{ij}^{xyz}(t) &= u_{ij}^{xyz}(0) + f(\mathbf{w}_{ij}^{r}u_{ij}^{xyz}(t-1)) + b_{ij} \\
        u_{ij}^{xyz}(0) &= f(\mathbf{w}^c_{(i-1)j} u_{(i-1)j}^{xyz})
    \end{split}
\end{equation}
where $u_{ij}^{xyz}(0)$ denotes the feed-forward output of convolutional layer, $u_{ij}^{xyz}(t-1)$ is the recurrent input of previous time, $\mathbf{w}_k^c$ and $\mathbf{w}^r_k$ are the vectorized feed-forward kernels and recurrent kernels, $b_{ij}$ is the bias for $j$th feature map in $i$th layer, $u_{ij}^{xyz}(0)$ is the output of convolutional output of previous layer and $f(\mathbf{w}^r_{ij}u_{ij}^{xyz}(t-1)))$ is induced by the recurrent connections. $f$ is defined as popular rectified linear unit (ReLU) function $f(x)=\max(0,x)$, followed by a local response normalization (LRN)

\begin{figure}[t]
    \centering
        \includegraphics[width=8.5cm]{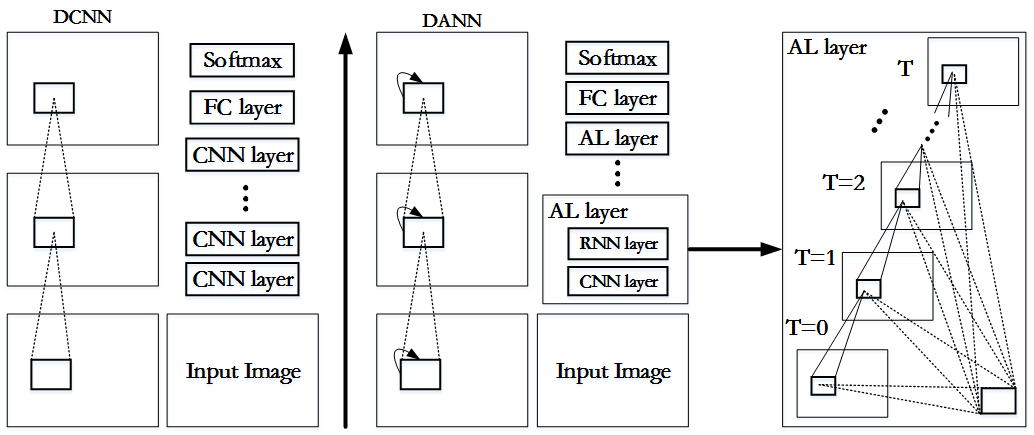}
    \caption{Comparison of DANN (right) and DCNN (left).}\label{fig:dann}
\end{figure}

\begin{equation}
    \mathrm{LRN}(u_{ij}^{xyz}) = \frac {u_{ij}^{xyz}}{{(1+\frac{\alpha}{\mathcal{L}}\sum_{k'=\max(0,k-\mathcal{L}/2)}^{\min(\mathcal{K},k+\mathcal{L}/2)}{(u_{ij}^{xyz})}^2)}^\beta}
\end{equation}
where $\mathcal{K}$ is the number of feature maps, $\alpha$ and $\beta$ are constants controlling the amplitude of normalization. The LRN forces the units in the same location to compete for high activities, which mimics the lateral inhibition in the cortex. In our experiments, LRN is found to consistently improve the accuracy, though slightly. Following \cite{krizhevsky2012imagenet}, $\alpha$ and $\beta$ are set to $0.001$ and $0.75$, respectively. $\mathcal{L}$ is set to $\mathcal{K}/8+1$.

Equation \ref{eq:al} describes the dynamic behavior of AL where contexts are involved after local features are extracted. Unfolding this layer for $\mathrm{T}$ time steps results in a feed-forward subnetwork of depth $\mathrm{T} + 1$ as shown in the right of Figure \ref{fig:dann}. While the recurrent input evolves over iterations, the feed-forward input remains the same in all iterations. When $\mathrm{T} = 0$ only the feed-forward input is present. The subnetwork has several paths from the input layer to the output layer. The longest path goes through all unfolded recurrent connections, while the shortest path goes through the feed-forward connection only. The effective RF of an AL unit in the feature maps of the previous layer expands when the iteration number increases. If both input and recurrent kernels in equation have square shapes in each feature map of size $\mathrm{L}_{\mathrm{feed}}$ and $\mathrm{L}_{\mathrm{rec}}$, then the effective RF of an AL unit is also square, whose side length is $\mathrm{L}_{\mathrm{feed}}+\mathrm{L}_{\mathrm{rec}}\times (\mathrm{T}+1)$.

\textbf{Advantages over DCNNs.} The recurrent connections in DANN provide three major advantages compared with popular DCNNs used for move prediction \cite{clark2015training,maddison2015move,tian2016better,silver2016mastering}. First, they enable every unit to incorporate contexts in an arbitrarily large region in the current layer, which is particular suitable in the game of Go as the input signal is very small where the contexts are essential. As the time steps increase, the state of every unit is influenced by other units in a larger and larger neighborhood in the current layer. In consequence, the size of regions that each unit can watch in the input space also increases. In standard convolutional layers, the size of effective RFs of the units in the current layer is fixed, and watching a larger region is only possible for units in higher layers. But unfortunately the context seen by higher-level units cannot influence the states of the units in the current layer without top-down connections. Second, the recurrent connections increase the network depth while keeping the number of adjustable parameters constant by weight sharing. Specially, stacking higher layers consume more parameters while AL uses only additional constant parameters compared to standard convolutional layer. This is consistent with the trend of modern deep architectures, i.e., going deeper with relatively small number of parameters \cite{simonyan2014very,szegedy2015going}. Note that simply increasing the depth of CNN by sharing weights between layers can result in the same depth and the same number parameters as DANN. We have tried such a model which leads to a lower performance. The third advantage is the time-unfolded manner in Figure \ref{fig:dann}, which is actually a CNN with multiple paths between the input layer to the output layer to facilitate the learning procedure. On one hand, the existence of longer paths makes it possible for the model to learn highly complex features. On the other hand, the existence of shorter paths may help gradient of backpropagation during training. Multi-path is also used in \cite{lee2015deeply,szegedy2015going}, but extra objective functions are used in hidden layers to alleviate the difficulty in training deep networks, which are not used in DANN.

\textbf{Overall architecture.}\label{sec:dann:overall} The overall architecture of our DANN has $6$ ALs with $64$, $128$, $256$, $256$, $512$ and $512$ kernels, followed by $2$ fully connected (FC) layers of size $1024$ each. We use $3 \times 3$ kernel for convolutional layer and recurrent layers of all 6 ALs. After each AL, the network includes a ReLU activation. We use max pooling kernels of $2 \times 2$ size. All of these convolutional layers and recurrent layers are applied with appropriate padding and stride. FC layers are followed by a ReLU and a softmax, which outputs the probabilities of Go board and illegal points are set $0$.

\subsection{Long-Term Evaluation of Candidates}

DANN provides a probability distribution of next move candidates give a situation. We further enhance this model by evaluating these candidates in a long-term consideration since predicting only the immediate next move limits the information received by lower layers \cite{tian2016better}. Besides, many situations in intensive battle or {\it{capture chase}} is far beyond fair evaluation and need to be accurately judged when local variation is settled. We aim to avoiding shortsighted moves. There are some works such as \cite{littman1994markov} that consider playing games as a sequential decision process of a goal-directed agent interacting with visual environment. We extend this idea to evaluate candidates in a similar manner. We calculate the cumulative rewards of each candidate with several future interactions. Combining previous probabilities criterion, we obtain a final score and determine the optimal action.

\begin{figure}[t]
  \centering
    \includegraphics[width=8.5cm]{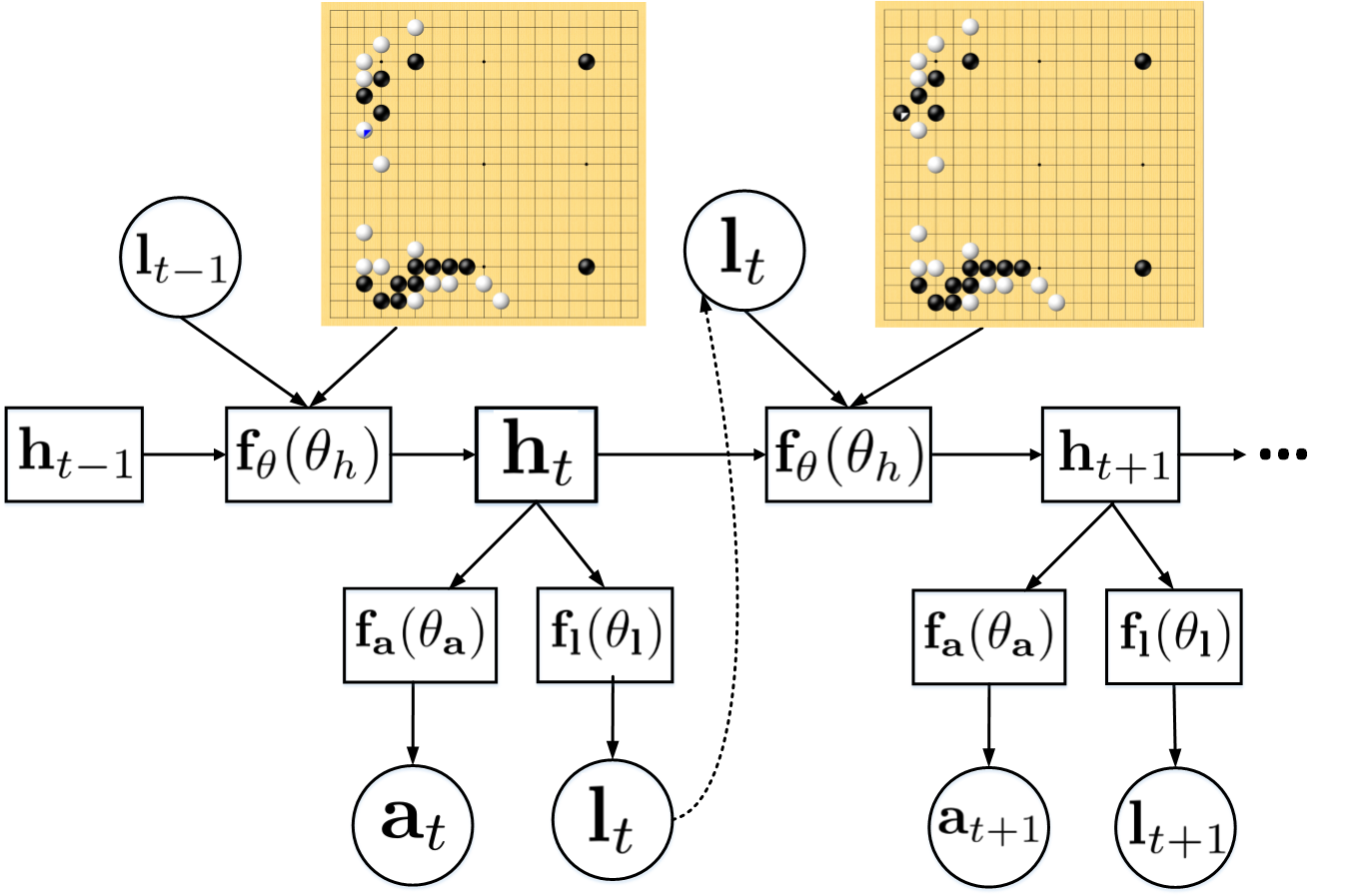}
  \caption{A recurrent model for long-term evaluation.}\label{fig:long}
\end{figure}

\textbf{Recurrent model and internal state.} Figure \ref{fig:long} shows our model structure, which is build a agent around a RNN. To avoiding blind search space like MCTS, the agent observes the environment only via a bandwidth-limited sensor, i.e. it never senses the full board. It may extract information only in a local region around candidates. The goal of our model is to provide a reliable evaluation of each candidate and assist the final choice. The agent maintains an internal state which summarizes information extracted from past observations. It encodes the agent's knowledge of the environment and is instrumental to deciding how to act and where to deploy the next action. This internal state is formed by the hidden units $\mathbf{h}_t$ of the recurrent neural network and updated over time by the core network $\mathbf{h}_t=\mathbf{f}_\mathbf{h}(\mathbf{h}_{t-1},\mathbf{l}_{t-1};\theta_\mathbf{h})$.

\textbf{Action and reward.} At each step, the agent performs two actions. It decides how to deploy its sensor via the sensor control $\mathbf{l}_t$, and an action $\mathbf{a}_t$ which might affect the state of the environment. The location are chosen stochastically from a distribution parameterized by the location network $\mathbf{l}_t \sim p(\cdot|\mathbf{f}_\mathbf{l}(\mathbf{h}_t;\theta_\mathbf{l})$. The action is similarly drawn from a distribution conditioned on a second network output at $\mathbf{a}_t \sim p(\cdot|\mathbf{f}_\mathbf{a}(\mathbf{h}_t;\theta_\mathbf{a})$. Finally, our model can also be augmented with an additional action that decides when it will stop when local fights are settled. After executing an action the agent receives a new visual observation and a reward signal $r$. The goal of the agent is to maximize the sum of the reward signal $\mathcal{R} = \sum_{t=1}^{T} r_t$. The above setup is a special instance of partially observable Markov decision process (POMDP), where the true state of whole board is unobserved.

\textbf{Training.} The policy of the agent, possibly in combination with the dynamics of interactions, induces a distribution over possible interaction sequences and we aim to maximize the reward under the distribution of
\begin{equation}\label{eq:aim}
\mathcal{J}(\theta)=\mathbb{E}_{p(s_{1:T};\theta)}[\sum_{t=1}^{T}r_t]=\mathbb{E}_{p(s_{1:T};\theta)}[\mathcal{R}]
\end{equation}

Maximizing $\mathcal{J}$ exactly is difficult because it involves an expectation over interaction sequences which may in turn involve unknown environment dynamics. Viewing the problem as a POMDP problem, however, allows us to bring techniques from the RL literature to bear. As shown in \cite{williams1992simple} a sample approximation to the gradient is given by
\begin{equation}\label{eq:rein}
        \begin{aligned}
        \nabla_\theta \mathcal{J}&=\sum_{t=1}^{T} \mathbb{E}_p(s_{1:T};\theta) [\nabla_\theta \log \pi (u_t|s_{1:t};\theta)\mathcal{R}] \\
        &\approx \frac {1}{M} \sum_{i=1}^{M} {\sum_{t=1}^{T} \nabla_\theta \log \pi (u_t^i|s_{1:t}^i;\theta)\mathcal{R}^i} \\
        \end{aligned}
\end{equation}
where $s^{i,}$s are the interaction sequences obtained by running the current agent $\pi_\theta$ for $i = 1 \cdots M$ episodes. The learning rule of Equation \ref{eq:rein} is also known as the REINFORCE rule, and it involves running the agent with its current policy to obtain samples of interaction sequences $s_{1:T}$ and then adjusting the parameters $\theta$ such that the log-probability of chosen actions that have led to high cumulative reward is increased, while that of actions having produced low reward is decreased. $\nabla_\theta \log \pi (u^i_t|s^i_{1:t};\theta)$ in Equation \ref{eq:rein} is just the gradient of the RNN and can be computed by standard backpropagation \cite{wierstra2007solving}.

\textbf{Variance reduction.} Equation \ref{eq:rein} provides us with an unbiased estimate of the gradient but it may have high variance. It is common to consider a gradient estimate of the form
\begin{equation}\label{eq:grad}
    \frac {1}{M} \sum_{i=1}^M \sum_{t=1}^T \nabla_\theta \log \pi (u^i_t|s^i_{1:t};\theta)(\mathcal{R}_t^i - b_t)
\end{equation}
where $\mathcal{R}_t^i = \sum_{t^{'}=1}^T r^i_{t^{'}}$ is the cumulative reward following the execution of action $u^i_t$, and $b_t$ is a baseline that depends on $s^i_{1:t}$ but not on the action $u^i_t$ itself. This estimate is equal to Equation \ref{eq:rein} in expectation but may have lower variance. We select the value function of baseline following \cite{sutton1999policy} in the form of $b_t = \mathbb{E}_\pi [R_t]$. We use this type of baseline and learn it by reducing the squared error between $\mathcal{R}_t^{i,}$ and $b_t$.

\textbf{Final score.} We define the final score for each candidate using the criteria of both DANN and long-term evaluation. We select the action of the highest score of $\mathcal{S} = p\times \mathbb{E}[\mathcal{R}]$ as the final choice of our system, where $p$ is the probability produced by the softmax layer of DANN.

\section{Experiments}\label{sec:eva}

\subsection{Setup}

\textbf{Datasets.} The first dataset we used is GoGoD (2015 winter version). The dataset consists of $82,609$ historical and modern games. We limited our experiments to a subset of games that satisfied the following criteria: $19\times 19$ board, modern (played after $1950$), ``standard'' komi (komi $\in \{5.5, 6.5, 7.5\}$), and no handicap stones. We did not distinguish between rulesets (most games followed Chinese or Japanese rules). Our criteria produced a training set of around $70,000$ games. We did not use popular KGS dataset because it consists of more games by lower {\it{dan}} players. The average level is approximately $5$ {\it{dan}}.

Besides, we have collected a new professional Go dataset (PGD) consisting of $253,233$ professional records, which exceeds GoGoD and KGS in both quantity and playing strength. PGD was parsed from non-profit web sites. All records are saved as widely used smart go file (SGF), named as \texttt{DT\_EV\_PW\_WR\_PB\_BR\_RE.sgf}, where \texttt{DT,EV,PW,WR,PB,BR} and \texttt{RE} represent date, tournament type, black player name, black playing strength, white player name, white playing strength and result.

\begin{table}[t]
    \footnotesize
    \centering
    \caption{Input feature channels for DANN.}\label{tab:feature}
    \begin{tabular}{p{1.89cm}p{0.1cm}p{5.9cm}}
          \toprule
          Feature & $\#$ & Description \\
          \midrule
          Ladder capture    & 1 & Whether point is a successful ladder capture \\
          Ladder escape     & 1 & Whether point is a successful ladder escape \\
          Sensibleness      & 1 & Whether point is legal and does not fill eyes \\
          Legality          & 1 & Whether point is legal for current player \\
          Player color      & 1 & Whether current player is black \\
          Zeros             & 1 & A constant plane filled with 0 \\
          Stone color       & 3 & Player stone/opponent stone/empty \\
          Liberties         & 4 & Number of liberties (empty adjacent points) \\
          Liberties$*$      & 6 & Number of liberties (after this move) \\
          Turn since        & 6 & Number of liberties (after this move) \\
          Capture size      & 8 & How many opponent stones would be captured \\
          Self-atari size   & 8 & How many own stones would be captured \\
          \bottomrule
          \hline
    \end{tabular}
\end{table}

\textbf{Feature channels.} The features that we used come directly from the raw representation of the game rules (stones, liberties, captures, legality, turns since) as in Table \ref{tab:feature}. Many of the features are split into multiple planes of binary values, for example in the case of liberties there are separate binary features representing whether each intersection has $1$ liberty, $2$ liberties, $3$ liberties, $>=4$ liberties.

\textbf{Implementation details.} The major implementations of DANN including convolutional layers, recurrent layers and optimizations are derived from Torch7 toolbox \cite{collobert2011torch7}. We use SGD applied to mini-batches with negative log likelihood criterion. The size of mini-batch is set $200$. Training is performed by minimizing the cross-entropy loss function using the backpropagation through time (BPTT) algorithm \cite{werbos1990backpropagation}. This is equivalent to using the standard BP algorithm on the time-unfolded network. The final gradient of a shared weight is the sum of its gradients over all time steps. The initial learning rate for networks learned from scratch is $3 \times {10}^{-3}$ and it is $3 \times {10}^{-4}$ for networks fine-tuned from pre-trained models. The momentum is set to $0.9$ and weight decay is initialized with $5 \times {10}^{-3}$ and reduced by ${10}^{-1}$ factor at every decrease of the learning rate.
\begin{table*}[t]\small
        \centering
        \caption{Performance comparison (top-1) with different configurations of DANN on GoGoD and PGD datasets.}\label{tab:model}
        \begin{tabular}{|m{1.9cm}|c|c|m{1.6cm}|c|c|}
                \hline
                    Architecture        & GoGoD & PGD   &   Architecture    & GoGoD     & PGD \\ \hline \hline
                    B\_6C\_2FC          & 32.2\%  & 37.2\%  &   2AL\_4C\_2FC    & 35.0\%      & 39.3\%\\
                    AL\_5C\_2FC         & 40.5\%  & 42.5\%  &   3AL\_3C\_2FC    & 42.7\%      & 43.1\%\\
                    C\_AL\_4C\_2FC      & 41.1\%  & 41.1\%  &   4AL\_2C\_2FC    & 47.4\%      & 42.2\%\\
                    2C\_AL\_3C\_2FC     & 45.6\%  & 44.8\%  &   5AL\_C\_2FC     & 47.2\%      & 46.4\%\\
                    3C\_AL\_2C\_2FC     & 47.4\%  & 43.0\%  &   6AL\_2FC        & 53.5\%      & 51.8\%\\
                    4C\_AL\_C\_2FC      & 49.4\%  & 46.6\%  &                   &           &\\
                    5C\_AL\_2FC         & 51.9\%  & 49.3\%  &                   &           &\\
                \hline
        \end{tabular}\label{tab:model:al} \ \ \ \ \ \
        \begin{tabular}{|m{2.2cm}|c|c|}
                \hline
                    Architecture        & GoGoD & PGD \\ \hline \hline
                    5AL\_2FC, $T$ = 2   & 46.9\%  & 42.3\% \\
                    5AL\_2FC, $T$ = 3   & 48.2\%  & 47.1\% \\
                    5AL\_2FC, $T$ = 4   & 52.3\%  & 45.8\% \\
                    5AL\_2FC, $T$ = 5   & 55.0\%  & 51.6\% \\ \hline    \hline
                    6AL\_2FC, $T$ = 2   & 46.4\%  & 47.2\% \\
                    6AL\_2FC, $T$ = 3   & 51.1\%  & 51.6\%\\
                    6AL\_2FC, $T$ = 4   & 55.3\%  & 49.4\%\\
                    6AL\_2FC, $T$ = 5   & 57.7\%  & 53.8\% \\
                \hline
        \end{tabular}\label{tab:model:t}
\end{table*}

\subsection{Move Prediction}

We first evaluate different configurations of DANN. Then we compare our best DANN model with DCNN-based methods. Finally, we study the impact of long-term evaluation and report the overall performance on move prediction.

\textbf{Model investigation of DANN.} There are two crucial configurations for DANN model. The first one is the AL setting including its order and number. The other one is the unfolding time $T$ in recurrent layers. Comparison details are reported in Table \ref{tab:model}, where B\_6C\_2FC is a baseline composed of similar configuration with DANN but using standard convolutional layers instead of ALs. The first column of left table in Table \ref{tab:model} has only one AL layer and the accuracy comparison demonstrates the benefits of inserting AL in advance. We attribute it to the context mining of lower features. The fourth column of left table in Table \ref{tab:model} shows the performance increases as the number of AL increases, which verifies the effectiveness of inserting recurrent layer. Specifically, the order of AL can contribute a performance gain up to $11\%$ which indicates that mining contexts of lower layer is beneficial for playing Go. Right table in Table \ref{tab:model} uses 5AL\_2FC and 6AL\_2FC to study the impact of $T$ and the results prove larger $T$ leads to better performance in most cases. Given such results we use our best DANN model 6AL\_2FC in the following experiments.

\begin{table*}[t]\footnotesize
    \centering
    \caption{Win rate comparison against open source engines between our system and previous work.}\label{tab:win rate}
        \begin{tabular}{|c|c|c|c|c|c|c|}
            \hline
                                                                &   GnuGo           &   MoGo 10k   &    Pachi 10k      &   Pachi 100k      &   Fuego 10k       &   Fuego 100k      \\    \hline  \hline
            8-layer-DCNN + MCTS \cite{clark2015training}   &    91.0\%         &   -       &    -              &     -             &     14.0\%        &       14.0\%        \\
            12-layer-DCNN + MCTS  \cite{maddison2015move}  &    97.2\%         &   45.9$\%$&   47.4\%          &   11.0\%          &     23.3\%        &   12.5\%            \\
            12-layer-DCNN + MCTS \cite{tian2016better}     &    100$\pm0.0$\%  &   -       &   94.3$\pm1.7$\%  &   72.69$\pm1.9$\% &    93.2$\pm1.5$\% &   89.7$\pm2.1$\%    \\
            6-layer-DANN + LTE (Ours)                       &    100$\pm0.0$\%  &   72.5$\pm1.8\%$&   83.1$\pm1.4$\%  &   65.3$\pm1.6$\%  &    82.6$\pm1.2$\% &   76.5$\pm1.6$\%    \\    \hline
        \end{tabular}
\end{table*}

\begin{figure}[t]
    \begin{tabular}{c}
        \includegraphics[width=8.6cm]{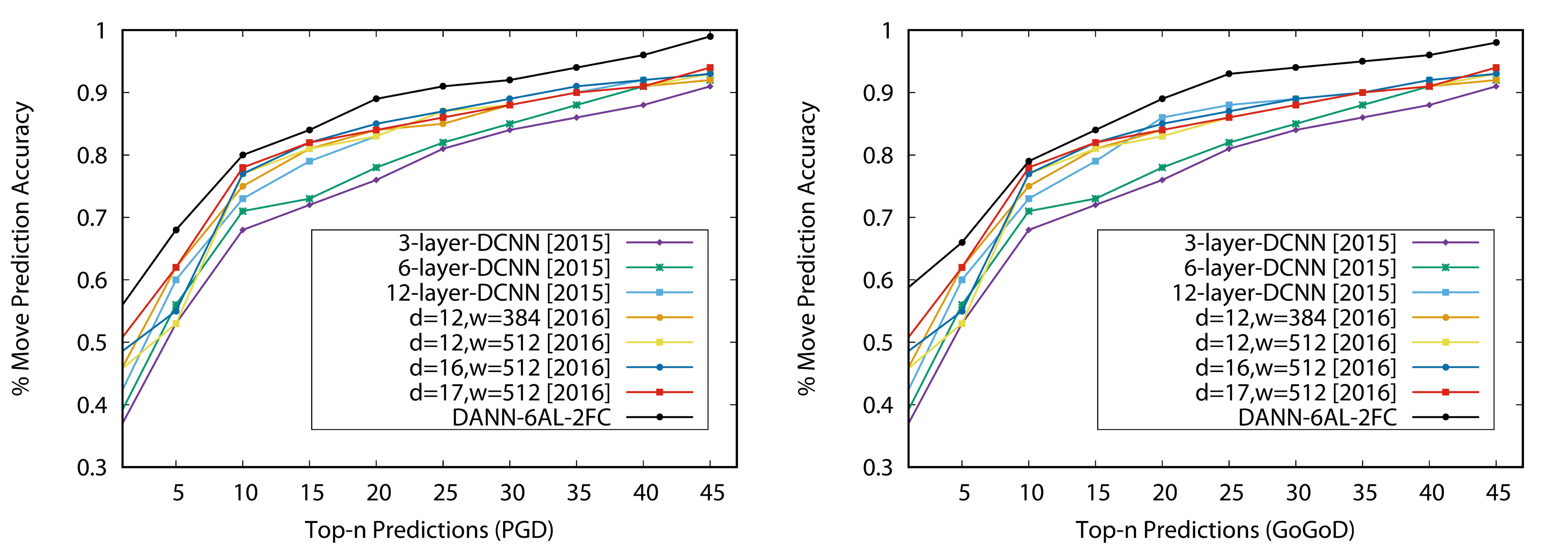}
    \end{tabular}
    \caption{Top-n comparison of our best DANN model and DCNN-based methods on GoGoD and PGD datasets.}\label{fig:exp:comp}
\end{figure}

\textbf{Comparison with DCNN-based methods.} Figure \ref{fig:exp:comp} reports the performance comparison of our best DANN model and related approaches using pure DCNN. Following \cite{maddison2015move} we evaluate the accuracy that the correct move is within the network¡¯s $n$ most confident predictions. As Figure \ref{fig:exp:comp} shows, our model consistently outperform two recent approaches \cite{maddison2015move,tian2016better} using pure DCNN on two datasets. Also, note that our architecture consume less layers where the parameters are also saved .

\begin{figure}[h]
    \begin{tabular}{c}
        \includegraphics[width=8.6cm]{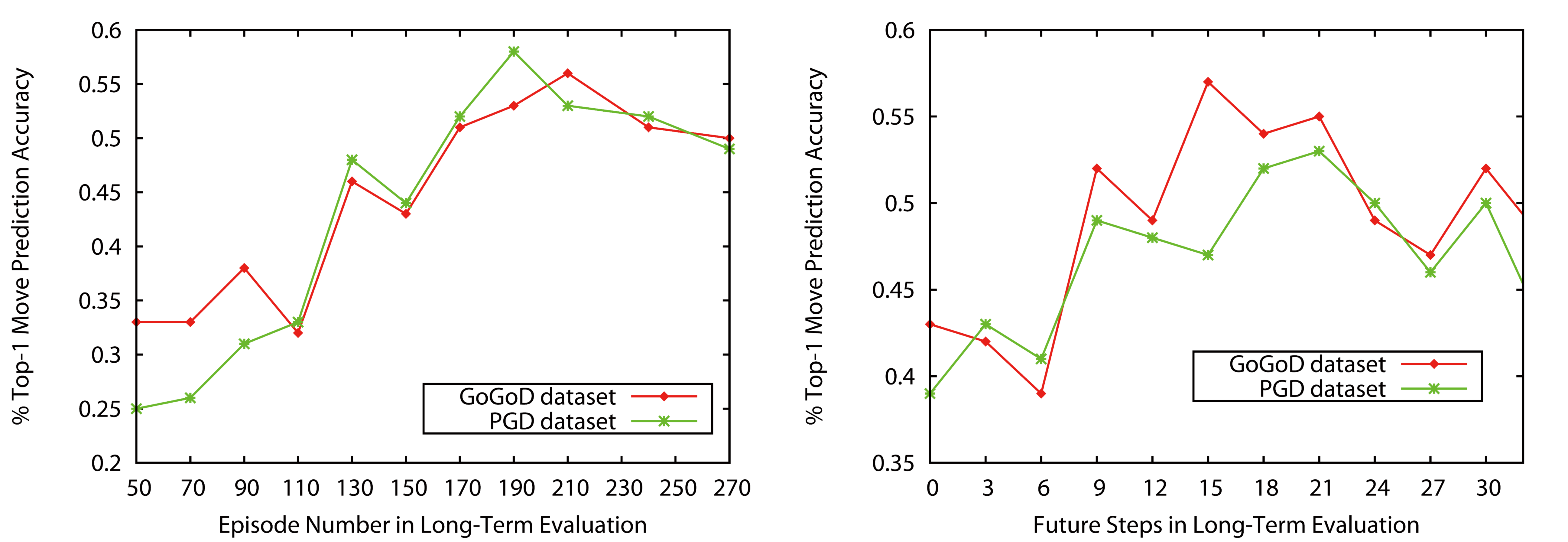}
    \end{tabular}
    \caption{Impact of future steps and episode number in long-term evaluation on GoGoD and PGD datasets.}\label{fig:exp:long}
\end{figure}

\textbf{Combining long-term evaluation.} Next we examine the influence of our long-term evaluation (LTE). We focus mainly on the future step that is used to achieve the expectation of reward, and the episode number when solving Equation \ref{eq:rein}. We combine our best DANN model with LTE on Top-1 move prediction accuracy. Table \ref{fig:exp:long} demonstrates the details. As can be seen, the best performance is achieved around $15$ to $21$ steps. As for the episode, LTE often converges after around $200$ episodes. Using the optimal setting of both part, the overall top-1 accuracy can be obtained at $61\%$ and $56\%$ on GoGod and PGD datasets, respectively.

\subsection{Playing Strength}

Finally, we evaluate the overall playing strength of our system by playing against several publicly available benchmark programs. All programs were played at the strongest available settings, and a fixed number of rollouts per move. We used GnuGo 3.8 level 10, MoGo \cite{gelly2007combining}, Pachi 11.99 (Genjo-devel) with the pattern files, and Fuego 1.1 throughout our experiments. For each setting, $3$ groups of $100$ games were played. We report the average win rate and standard deviation computed from group averages. All the game experiments mentioned in this paper used komi $7.5$ and Chinese rules. Pondering (keep searching when the opponent is thinking) in Pachi and Fuego are on. As Table \ref{tab:win rate} shows, our system outperform most MCTS-based Go programs. Also, the win rate of our approach is higher than that of previous works except \cite{tian2016better}.

\section{Conclusion}

In this work, we have proposed a computer Go system based on a novel deep alternative neural networks (DANN) and long-term evaluation (LTE). We also public a new dataset consisting of around $25$k professional records. On two datasets, we showed that DANN can predict the next move made by Go professionals with an accuracy that substantially exceeds previous deep convolutional neural network (DCNN) methods. LTE strategy can further enhance the quality of candidates selection, by combining the influence of future interaction instead of immediate reward. Without brute simulation of possible interaction in a large and deep search space, our system is able to outperform most MCTS-based open source Go programs.

Future work mainly includes the improvement of DANN structure for move prediction and more reliable LTE implementation. Advance techniques in computer vision community such as residual networks may help DANN obtain further improvement. As for LTE, domain knowledge of Go will be attempted to provide a more reliable estimation of next move candidates.

\subsection{Acknowledgement}

This work was supported by Shenzhen Peacock Plan (20130408-183003656).

\bibliographystyle{aaai}
\bibliography{wang}
\end{document}